\documentclass[10pt,twocolumn,letterpaper]{article}

\usepackage{cvpr}              % To produce the CAMERA-READY version
\usepackage{multirow}

\definecolor{cvprblue}{rgb}{0.21,0.49,0.74}
\usepackage[pagebackref,breaklinks,colorlinks,allcolors=cvprblue]{hyperref}

\def\confName{CVPR}
\def\confYear{2026}

\title{Spatio-Temporal Conditional Denoising Transformer for Modality-Missing RGBT Tracking}

\author{Andong Lu$^{1}$ \quad Ziyi Zha$^{1}$ \quad Jiandong Jin$^{1}$ \quad Shihao Li$^{2}$ \quad Chenglong Li$^{2\dagger}$ \quad Jin Tang$^{1}$ \quad Bin Luo$^{1}$\\
$^{1}$ School of Computer Science and Technology, Anhui University, China\\
$^{2}$ School of Artificial Intelligence, Anhui University, China\\
{\tt\small \{adlu\_ah, ziyi.zha, jdjinahu, shli0603, lcl1314\}@foxmail.com, \{tangjin, luobin\}@ahu.edu.cn}
}

\begin{document}
\maketitle
\let\thefootnote=\relax\footnotemark\footnotetext{$^{\dagger}$Corresponding author.}
\begin{abstract}
% Missing modalities are an unavoidable challenge in practical RGB-Thermal (RGBT) tracking, and effectively generating or adapting to missing modalities is crucial for robust tracking.
%
% Current methods typically rely on inferring missing information from the available modality, or on introducing multiple expert networks to handle different input conditions. However, these strategies are either restricted by single-modality limitations or suffer from redundancy and inconsistent representation learning.
%
% modality generation and enhancement are governed by a noise-conditioned interaction mechanism: given the current-frame features of the available modality and the historical context of the complementary one, the module predicts target-modality features for completion or enrichment. By modulating the injected noise intensity, the network seamlessly transitions between generation and enhancement behaviors, achieving unified, efficient, and context-aware multimodal representation learning.

% Modality-missing RGB-Thermal (RGBT) tracking remains a critical challenge in real-world scenarios, where the absence of one modality often leads to incomplete and unstable feature representations.
%
% Existing methods attempt to recover missing information from available cues, yet single-modality cues are insufficient to reliably reconstruct due to significant cross-modal differences.
%
Missing modalities in RGBT tracking often lead to incomplete and unstable multimodal feature representations that greatly degrade the performance. Existing methods typically attempt to recover missing modalities from available ones, but the quality of data generated in challenging scenarios might be unsatisfactory. In addition, current approaches exhibit limited flexibility in processing both missing and complete data.
To overcome these limitations, we propose a Spatio-temporal Conditional Denoising Transformer (SCDT), which integrates the spatial cues and the temporal context to adaptively perform information reconstruction of missing modalities and feature enhancement of weak modalities in a unified framework, for robust modality-missing RGBT tracking. 
In particular, SCDT leverages the short-term temporal cues from recent historical frames to capture the fine-grained temporal correlations and the long-term temporal cues encoding modality evolution to capture the global context. By jointly exploiting long short-term temporal contexts as the conditions, SCDT progressively guides noisy features of available modalities to learn reliable and temporally consistent multimodal representations. 
Furthermore, SCDT introduces a noise-modulated adaptation mechanism that dynamically adjusts its behavior according to the modal availability, enabling a single framework to unify feature learning under both modality-missing and complete scenarios without changing the architecture or parameters.
Extensive experiments on three public benchmark datasets demonstrate that our method consistently outperforms state-of-the-art methods. The code is available \href{https://github.com/Multi-Modality-Tracking/SCDT-CVPR2026}{here}.

%\url{https://github.com/Multi-Modality-Tracking/SCDT-CVPR2026}}.
% % This unified modeling enables efficient, consistent, and adaptive multimodal representation learning under varying modality availability.
% This design enables the model to dynamically reconstructs reliable representations when modalities are missing, and enriches feature diversity when all modalities are available without switching architectures or parameters. 
% Furthermore, by executing different constraints, Unit effectively adapts to both missing and complete inputs within a unified model, regardless of modal availability. Specifically, a strict feature-level constraint helps fine reputation when modality missing occurs, while a mild statistical constraint is introduced to encourage diversity presentation while preserving semantic consistency when both modalities are available. 
\end{abstract}

% We formulate our multimodal fusion as a conditional denoising diffusion process. By injecting varying levels of noise into the available modality of the current frame, DiffusionBridge adaptively modulates its generation objective within a unified diffusion model.     
\section{Introduction}
\label{sec:intro}

Object tracking aims to continuously localize a target continuously across frames, which remains a fundamental yet challenging task in computer vision. Among multimodal tracking paradigms, RGB-Thermal (RGBT) tracking has gained increasing attention for its robustness under adverse conditions~\cite{2021MANet++,Zhang_CVPR22_VTUAV,SDSTrack,BAT2024,AINet2025,li2025cadtrack}. The RGB modality provides detailed appearance and semantic information under normal illumination, whereas the thermal modality offers complementary radiometric cues that are stable in low-light or obscured scenes. By leveraging these complementary advantages, RGBT tracking has shown great potential in safety-critical tasks such as nighttime surveillance~\cite{lu2025nighttime}, search and rescue~\cite{cruz2024thermal}, and autonomous driving~\cite{MIR-2025-02-072}.

\begin{figure}[t]
    \centering
    \includegraphics[width=1.0\linewidth]{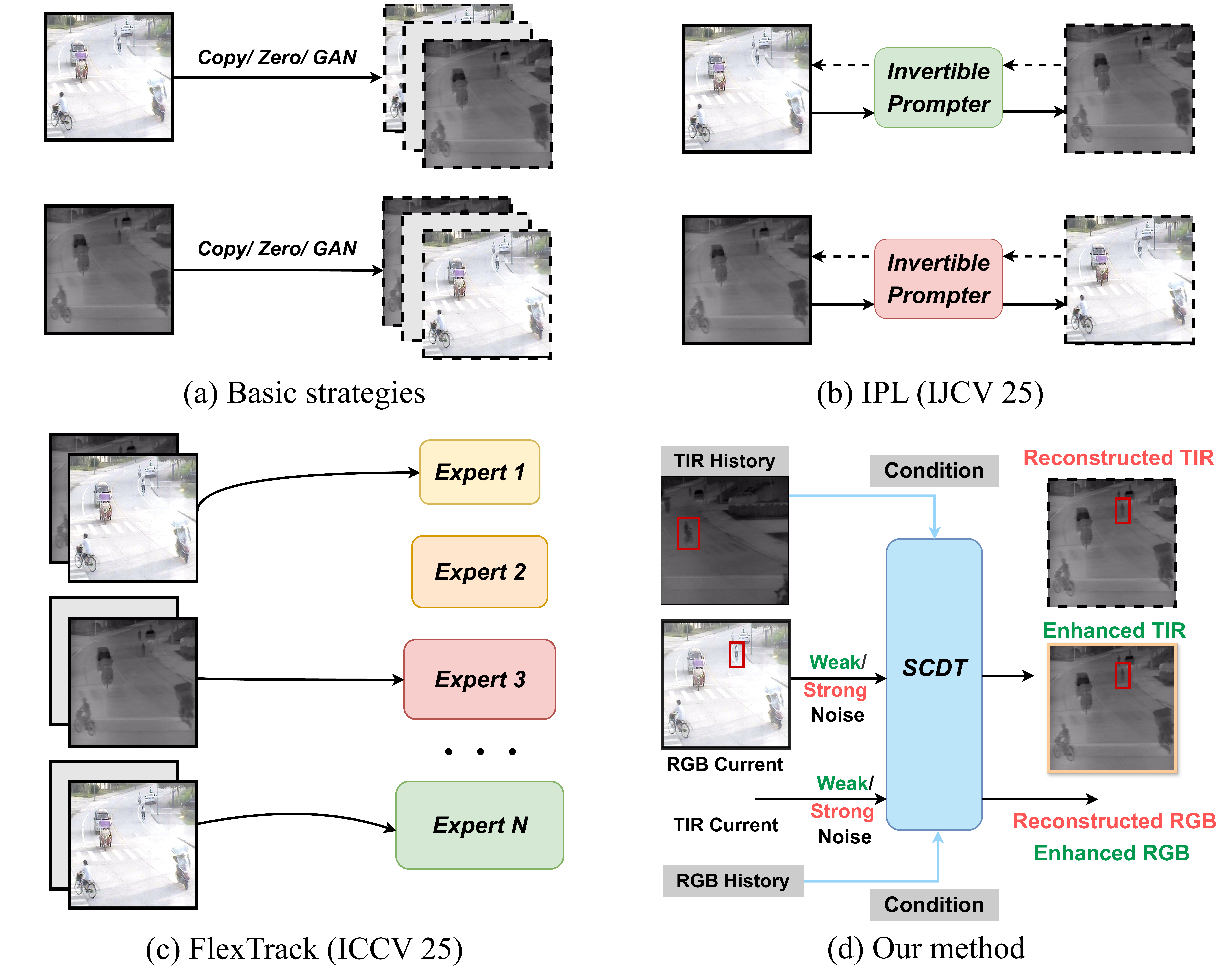}
     \caption{Comparison of the existing method of RGBT tracking when facing missing scenerios. (a) Basic strategies. (b) Invertible prompt learning. (c) Mixture of Experts. (d) Our method, uses a unified Spatio-temporal Denoising module to handle both complete and missing scenerios without changing architectures.
     }
    \label{motivation_fig}
\end{figure}

% Unlike RGB-Depth or RGB-Event setups, RGBT sensors operate under fundamentally different imaging principles, making them more prone to inconsistent availability and asynchronous degradation.

However, the practical deployment of RGBT systems is often hindered by modality-missing issues caused by sensor misalignment, occlusion, or hardware malfunction. When one modality becomes unavailable, the feature representations learned by network become incomplete or unstable, which significantly undermines tracking robustness. This dynamic modality missing poses a unique challenge: the tracker must not only reconstruct missing information but also adaptively exploit the temporal and cross-modal context to maintain coherence over time.
Existing approaches mainly tackle this challenge through modality reconstruction or modality-specific fusion strategies, as illustrated in Fig.~\ref{motivation_fig}. For instance, Lu et al.~\cite{lu2025modality} attempt to generate the missing modality features from the available one, while Tan et al.~\cite{tan2025you} design mixture-of-experts or switch-based architectures to handle different modality configurations. Despite their progress, these methods face two fundamental limitations. First, they primarily rely on the current-frame cues of the available modality, neglecting the temporal correlations from historical frames that could provide valuable information about the missing modality. As a result, the reconstructed features tend to be spatially biased or temporally inconsistent. Second, their architectures are often scenario-dependent, requiring explicit switching or separate branches to process missing and complete cases, which limits scalability and leads to inefficiency.

To address these limitations, we propose the Spatio-temporal Conditional Denoising Transformer (SCDT), a unified framework designed to handle both missing and complete modality conditions within a single model, as shown in Fig.~\ref{motivation_fig}(d). SCDT reformulates multimodal feature reconstruction as a spatio-temporal conditional denoising process that integrates spatial cues from the current frame with temporal contexts derived from the historical frames of the missing modality. It leverages short-term temporal information from adjacent frames to capture fine-grained motion continuity and long-term temporal cues encoding modality evolution to provide stable global context. These complementary temporal conditions jointly guide the denoising process, enabling SCDT to progressively refine noisy inputs from available modalities into reliable and temporally coherent multimodal representations. This design effectively bridges modality discrepancies while maintaining consistency across time, leading to robust and stable tracking under modality missing.

In addition, SCDT introduces a noise-modulated adaptation mechanism that unifies modality reconstruction and enhancement within a single framework. By injecting noise of varying intensity into the available modalities, the model implicitly encodes distinct fusion objectives. Strong noise encourages reconstruction under missing conditions, while mild noise promotes feature enhancement under complete scenarios. Through this unified yet adaptive learning scheme, SCDT maintains consistent parameters and computation across all modality conditions, ensuring efficient and consistent feature learning without redundant computation. Our main contributions can be summarized as follows:

\begin{itemize}

\item We propose SCDT, a unified spatio-temporal conditional denoising framework that effectively handles both missing and complete modalities in RGBT tracking.

\item We design a dual temporal conditioning strategy that jointly models short-term and long-term temporal cues to achieve coherent and context-aware feature reconstruction.

\item We develop a noise-modulated adaptation mechanism that enables seamless adjustment between reconstruction and enhancement objectives while maintaining consistent parameters and computation across all modality conditions.

\item Extensive experiments on three challenging RGBT tracking benchmarks demonstrate that SCDT sets new state-of-the-art performance across diverse modality conditions, establishing a strong baseline for future modality-missing tracking research.
\end{itemize}

\section{Related Work}
\label{sec:formatting}
%-------------------------------------------------------------------------
\subsection{RGBT Tracking}

RGBT tracking leverages complementary cues from RGB and thermal modalities for robust localization under adverse conditions. Early studies primarily focus on modality-complete settings, emphasizing spatial correspondence to exploit cross-modal complementarity. Representative works include TBSI~\cite{li2024rgb}, which introduces cross-modal search region interaction through attention mechanism, and AINet~\cite{AINet2025}, which employs Mamba-based fusion to enhance inter-modal interactions.
% BAT工作替换成出具有显式交互模块的方案，1.TBSI的期刊拓展版本
Subsequent research incorporates temporal modeling to improve motion consistency~\cite{zhang2024exploring,hu2025exploiting}, but these approaches mainly capture intra-modal dependencies, overlooking cross-modal temporal relations that are crucial for robust fusion in dynamic environments.
When one modality becomes unavailable, the performance of these trackers drops sharply. Recent works such as IPL~\cite{lu2025modality} and FlexTrack~\cite{tan2025you} attempt to address modality-missing scenarios through generative reconstruction or adaptive expert routing. However, these designs remain limited by their reliance on spatial cues and lack of explicit temporal consistency modeling.
Our work addresses this gap by couple spatial and temporal information to enhance stability under modality-missing conditions.
%-------------------------------------------------------------------------

\subsection{Missing Modality Learning}
Handling incomplete modalities has become a key topic in multimodal learning, where missing or corrupted inputs are common due to sensor limitations or environmental factors. A variety of strategies have been explored across different domains such as visual-language understanding~\cite{sim2025can}, audio-visual learning~\cite{park2024learning}, and multimodal recognition~\cite{yun2024flex}. 
Existing approaches can be broadly categorized into feature reconstruction, which estimates missing representations from available modalities~\cite{woo2023towards}, and knowledge transfer, where full-modality models guide models operating under incomplete inputs~\cite{li2024toward}. In addition, prompt-based strategies have also been explored~\cite{xu2025distilled}, but uniform prompts often limit per-instance adaptability, which is critical in video tracking.
Despite their effectiveness, most existing approaches are designed for static or single-frame tasks and do not explicitly leverage temporal cues. Extending them to video-based tracking remains challenging, as temporal cues from historical frames are essential for compensating missing modalities. Our study builds upon this insight, emphasizing temporally informed feature reconstruction to improve reliability in modality-missing RGBT tracking.

\subsection{Diffusion Models}
Diffusion-based denoising models~\cite{ho2020denoising,song2020denoising,nichol2021improved} have emerged as a powerful generative framework that formulates data synthesis as an iterative denoising process, progressively refining random Gaussian noise into structured signals. Existing diffusion architectures can be broadly categorized into convolutional U-Net-based models and transformer-based variants. The former, including DALL-E 2~\cite{ramesh2022hierarchical}, Imagen~\cite{saharia2022photorealistic}, and Stable Diffusion~\cite{rombach2022high}, rely on hierarchical convolutional encoders and decoders, with latent diffusion further enhancing efficiency through VAE-based compression. The latter, known as Diffusion Transformers~\cite{peebles2023scalable}, replace convolutional backbones with transformer blocks that jointly encode time steps, noise levels, and conditional tokens, demonstrating strong scalability and generalization in vision generation tasks.
Recent studies have extended diffusion models to conditional and cross-modal settings~\cite{zhang2023adding,wu2024medsegdiff}, enabling controllable and multimodal generation. A few works explore diffusion for visual tracking~\cite{tang2024generative,xie2024diffusiontrack}, but most remain limited to frame-level generation and neglect temporal dependencies critical for tracking tasks.
Our approach reinterprets the denoising process as a spatio-temporal conditional reconstruction mechanism, enabling dynamic refinement of multimodal representations under both complete and missing modality conditions.

\section{Method}
\label{sec:formatting}

\begin{figure*}
  \centering
    %\fbox{\rule{0pt}{2in} \rule{.9\linewidth}{0pt}}
    \includegraphics[width=0.95\linewidth]{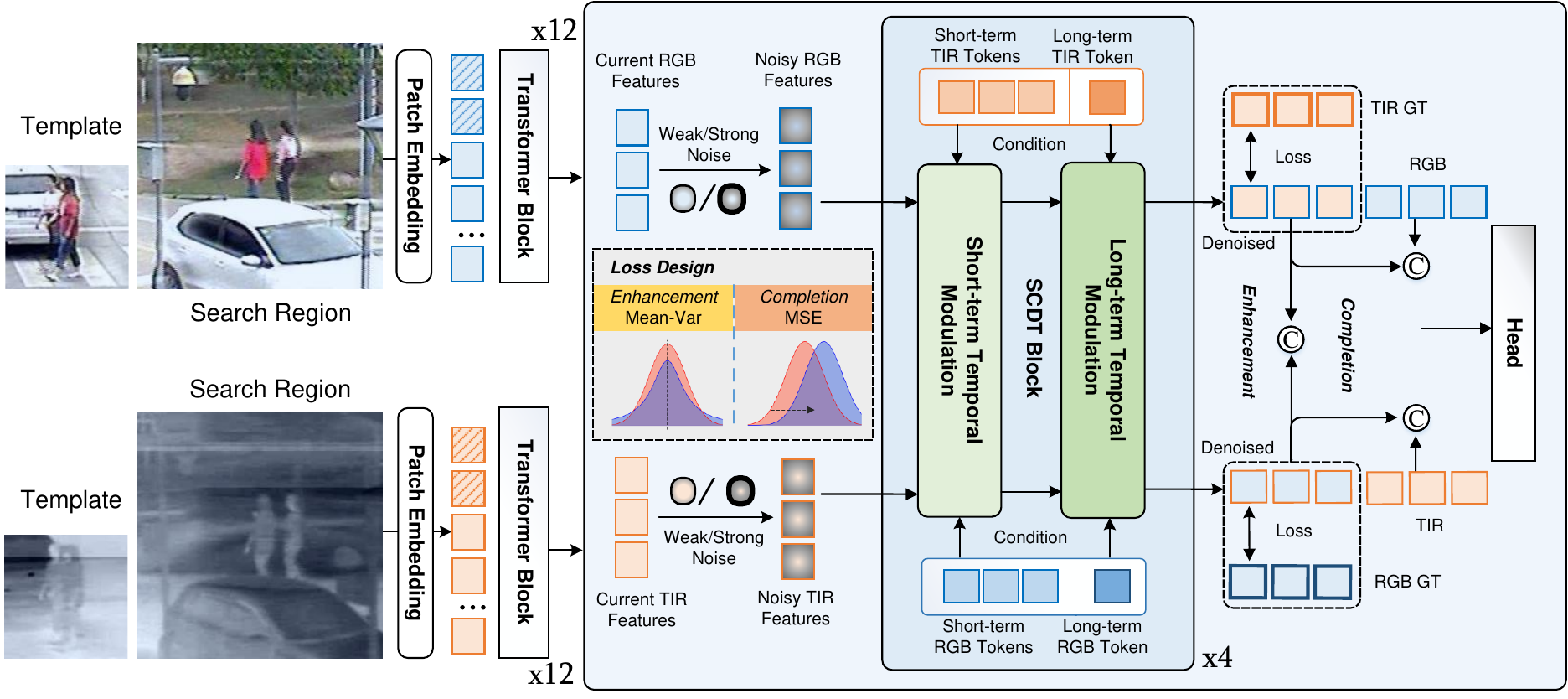}
    \caption{Overall architecture of the proposed SCDT. SCDT performs a noising-denoising process on available search frames, guided by historical frame features with short-term and long-term temporal dependencies. This enables simultaneous feature enhancement for complete modalities and reconstruction for missing modalities within a unified framework.}
    \label{fig:short-a}
\end{figure*}

%-------------------------------------------------------------------------
\subsection{Overview}
In this paper, we propose a Spatio-temporal Conditional Denoising Transformer (SCDT) for RGBT tracking, which unifies modality reconstruction and enhancement within a single generative framework, as illustrated in Fig.~\ref{fig:short-a}.
Given RGB and thermal video sequences, both modalities are independently processed by a shared ViT-B encoder to extract spatio-temporal representations from multiple template and search frames. The encoded features are then perturbed with adaptive Gaussian noise, forming inputs for the SCDT module.
The SCDT performs conditional denoising guided by spatial cues from the available modality and temporal cues from the complementary one. In modality-missing scenarios, SCDT restores the absent modality by leveraging spatial guidance from the available input and temporal information from historical frames. Under complete conditions, it also refines cross-modal representations to improve semantic alignment and temporal coherence.
Finally, the denoised features are fused and fed into the tracking head for bounding-box regression.
This unified framework enables robust adaptation across both complete and missing modality scenarios without altering the network architecture or parameters.

% The RGB and thermal (TIR) inputs are first processed by a shared transformer encoder, where each forward pass takes a sequence of three template frames and one search frame for each modality. After encoding, the search frame features from the RGB and TIR branches are denoted as \(F_{r},F_{t}  \in \mathbb{R}^{B,N_{s},C}\), which are subsequently fed into the SCDT module.
%
% The SCDT performs conditional denoising to reconstruct or refine multimodal representations. It receives spatial guidance from the available modality and temporal context from the complementary modality, enabling adaptive cross-modal generation. Under modality-missing conditions, SCDT performs unidirectional reconstruction; when both modalities are available, it conducts bidirectional enhancement. The network architecture and parameters remain identical in both cases. The restored or enhanced features are finally passed to the tracking head for bounding-box prediction.
%spatial cues from the current frame and 

%------------------------------------------------------------------------
\subsection{Spatio-temporal Conditional Denoising Transformer}

SCDT formulates multimodal fusion as a conditional generation process driven by spatio-temporal context. Instead of directly fusing heterogeneous features, SCDT learns to \emph{generate} modality representations conditioned on the available modality and temporal cues, enabling robust reconstruction under modality missing and adaptive cross-modality enhancement under modality complete input.

\noindent\textbf{Conditional Denoising Formulation.} Given the encoded feature $f_m \in \mathbb{R}^{B \times N \times C}$ of the available modality, we generate a noisy input for the denoiser:
\begin{equation}
    \tilde{f}_m = \sqrt{\bar{\alpha}}\, f_m + \sqrt{1 - \bar{\alpha}}\, \varepsilon, 
    \qquad \varepsilon \sim \mathcal{N}(0, \sigma^2 I),
\end{equation}
where the noise variance $\sigma^2$ is task-dependent: higher noise encourages reconstruction, while lower noise favors feature enhancement. The denoiser $D_\theta$ produces a refined feature conditioned on spatial and temporal cues:
\begin{equation}
    \hat{f} = D_\theta(\tilde{f}_m; c_s, c_t),
\end{equation}
where $c_s$ captures spatial information from the current frame, and $c_t$ integrates both short-term historical frames that are not missing and the long-term modality token. Implemented as a transformer, the denoiser fuses noisy feature tokens with condition tokens to generate semantically coherent outputs.

\noindent\textbf{Modality Feature Reconstruction.}  
When one modality (e.g., TIR) is missing, SCDT performs \emph{modality reconstruction}. We apply a stronger noise level to the available modality, forcing the denoiser to infer the missing modality semantics guided by spatio-temporal conditions:
\begin{equation}
    \hat{f}_{m'} = D_\theta(\tilde{f}_m; c_s, c_t).
\end{equation}
The reconstructed feature $\hat{f}_{m'}$ is concatenated with $f_m$ for downstream tracking. Training is supervised by a feature-level reconstruction objective:
\begin{equation}
    \mathcal{L}_{\text{recon}} = \| \hat{f}_{m'} - f_{m'} \|_2^2,
\end{equation}
which encourages accurate recovery of the missing modality in both spatial structure and semantic content.

\noindent\textbf{Modality Feature Enhancement.}  
When both modalities are available, SCDT follows the same conditional generation pathway under a weak-noise regime. Instead of enforcing pixel-wise fidelity, we encourage the model to generate an \emph{enhanced} feature $\hat{f}_m$ that remains close to the true distribution while offering improved discriminability. 
To this end, we align the first- and second-order statistics of the generated and real features:
\begin{equation}
    \mathcal{L}_{\text{align}} =
    \| \mu(\hat{f}_m) - \mu(f_m) \|_2^2 +
    \| \mathrm{Var}(\hat{f}_m) - \mathrm{Var}(f_m) \|_2^2,
\end{equation}
where $\mu(\cdot)$ and $\mathrm{Var}(\cdot)$ denote the mean and variance computed across spatial tokens. This statistical alignment encourages $\hat{f}_m$ to preserve the overall modality distribution while exploring discriminative directions beyond the deterministic reconstruction target, effectively enhancing feature expressiveness and robustness.

\noindent\textbf{Discussion.}
SCDT unifies reconstruction and enhancement within a single conditional generation framework. By adjusting the noise level and the training objective, it flexibly adapts to modality availability. Short-term and long-term temporal conditioning ensures spatially detailed and temporally coherent features, providing a principled solution for robust multimodal tracking under both missing and complete modality scenarios.

\begin{figure}
  \centering
    %\fbox{\rule{0pt}{2in} \rule{.9\linewidth}{0pt}}
    \includegraphics[width=0.8\linewidth]{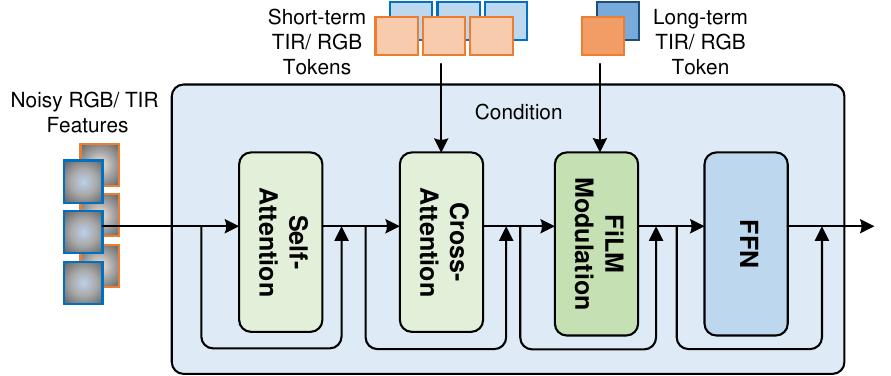}
    \caption{Detailed architecture of the Spatio-temporal Conditioned Denoiser Block.}
    \label{fig:short-b}
\end{figure}

\subsection{Spatio-temporal Conditioned Denoiser Block}
To effectively leverage temporal cues in feature refinement, each denoiser block in SCDT integrates short-term and long-term conditions in a complementary manner.

Given the noisy feature \(\tilde{f}_m\) of the available modality, short-term cues \(s_c\) are first fused via cross-attention:
% \begin{equation}
% f_m' = \tilde{f}_{m} + \text{CrossAttn}(\text{LN}(\tilde{f}_{m}), s_c, s_c),
% \end{equation}
\begin{equation}
f_m' = \tilde{f}_{m}^{SA} + \text{CrossAttn}(\tilde{f}_{m}^{SA}, s_c, s_c),
\end{equation}
where \(s_c\) represents short-term temporal tokens extracted from recent neighboring frames of the complementary modality that are not missing. This allows the denoiser to leverage locally aligned, frame-level information to recover or enhance target-relevant features and mitigate cross-modal misalignment.
Next, a global long-term token \(l_c\) modulates the features via a FiLM-style~\cite{perez2018film} scale and shift operation:
\begin{equation}
f_m'' = f_m' \odot \big(1 + \tanh(W_s l_c)\big) + \tanh(W_r l_c),
\end{equation}
where \(l_c\) encodes long-term temporal context across the sequence to stabilize features and suppress noisy activations, and \(W_s, W_r\) are learnable projections.
Finally, a feed-forward network produces the refined feature:
\begin{equation}
\hat{f}_m = f_m'' + \text{FFN}(\text{LN}(f_m'')).
\end{equation}
By applying short-term cross-attention and long-term FiLM modulation, the denoiser effectively balances local alignment and global coherence, enabling robust reconstruction or enhancement under varying modality availability.

\subsection{Loss Function}

After the denoising process, the enhanced or reconstructed features from both modalities are concatenated along the feature dimension and fed into a fully convolutional prediction head to generate the predicted bounding box. The overall training objective combines multiple loss terms:
\begin{equation}
\mathcal{L}_{\text{total}} =
\lambda_1 \mathcal{L}_{\text{recon}} +
\lambda_2 \mathcal{L}_{\text{align}} + 
\lambda_3 \mathcal{L}_{\text{track}},
\end{equation}
where $\mathcal{L}_{\text{recon}}$ encourages accurate reconstruction of missing modalities, $\mathcal{L}_{\text{align}}$ enforces alignment of feature statistics for modality enhancement, and $\mathcal{L}_{\text{track}}$ is the standard tracking loss following the settings in ODTrack~\cite{zheng2024odtrack}. The tracking loss weight is always set to $\lambda_3 = 1.0$. 

The weighting factors $\lambda_1$ and $\lambda_2$ are dynamically adjusted based on modality availability. For missing-modality scenarios, we set $\lambda_1 = 1.0$ and $\lambda_2 = 0.0$ to prioritize reconstruction; for complete-modality scenarios, we set $\lambda_1 = 0.0$ and $\lambda_2 = 1.0$ to emphasize feature enhancement. 
Sharing identical denoiser weights across both scenarios ensures a unified model capable of handling modality gaps seamlessly, maintaining robust tracking performance under diverse conditions.

% After the denoising process, enhanced or generated features of the two modality are concatenated in feature dimension and sent to the fully convolutional based prediction head, acquire the predicted box and calculate loss. The overall loss combines all objectives:
% \begin{equation}
% \mathcal{L}_{\text{total}} =
% \lambda_1 \mathcal{L}_{\text{r}} +
% % \lambda_2 \mathcal{L}_{\text{con}} +
% \lambda_2 \mathcal{L}_{\text{stat}} + 
% \lambda_{track} \mathcal{L}_{\text{track}}
% .
% \end{equation}
% The $\lambda_{track} \mathcal{L}_{\text{track}}$ follows the hyperparameter setting in OSTrack~\cite{ostrack}. For missing conditions, $\lambda_1$ is set to 1.0 and $\lambda_2$ is set to 0.0. For complete conditions, $\lambda_1$ is set to 0.0 and $\lambda_2$ is both set to 1.0.
% Identical parameters and denoiser weights ensures a unified model that performs consistently across both missing and complete modality scenarios, enablING the model to seamlessly handle modality gaps and maintain robust performance across diverse tracking conditions

%-------------------------------------------------------------------------

\section{Experiments}

\subsection{Impletation Details}
Our model is initialized using the pretrained model from ODTrack~\cite{zheng2024odtrack}, which is trained on four large-scale RGB tracking datasets. The template size is 128$\times$128 pixels and the search size is 256$\times$256 pixels, using the ViT-B backbone. The model is trained on 6 NVIDIA GeForce RTX 4090 GPUs with a total batch size of 24, using the AdamW optimizer. The learning rate is set to \(10^{-5}\) for the backbone and \(10^{-4}\) for the rest. For the LasHeR~\cite{li2021lasher}/LasHeR-missing~\cite{lu2025modality} and RGBT234~\cite{li2019rgb234}/RGBT234-missing~\cite{lu2025modality} datasets, the training lasts for 30 epochs, with each epoch contains 40000 samples. The weight decay is set to \(10^{-4}\) starting from epoch 24. For VTUAV~\cite{Zhang_CVPR22_VTUAV}/VTUAV-missing~\cite{lu2025modality} dataset, the training lasts for 5 epochs with 60000 samples of each epoch. The weight decay is set to \(10^{-4}\) after epoch 4. 
% During training, we introduce three patterns, including complete, RGB missing and TIR missing data. The rate of each pattern is set to equal.

\subsection{Evaluation Datasets and Protocol}

\noindent\textit{Complete RGBT Datasets.} 
We evaluate our method on three widely used RGBT tracking benchmarks.  
\textbf{LasHeR} contains 1,224 sequences with 1.47 million frames, spanning both surveillance and mobile scenarios. Its sequences have an average length of 600 frames and a maximum of 12,862 frames, with a resolution of 630 by 480 pixels.  
\textbf{RGBT234} includes 234 sequences totaling 116.7K frames, captured under various challenging conditions, with an average of 498 frames per sequence and a resolution of 630 by 460 pixels.  
\textbf{VTUAV} consists of 500 UAV-captured sequences with 1.7 million frames, featuring long sequences up to 27,213 frames, at 1920 by 1080 resolution, covering diverse aerial tracking scenarios.

\noindent\textit{Modality-Missing RGBT Datasets.} 
To evaluate tracking under modality degradation, we use high-quality modality-missing variants constructed from the original datasets.  
\textbf{LasHeR-Miss} is derived from the 245 sequences of LasHeR’s test set, containing 220.7K frames with 133.2K frames affected by missing modalities. Each sequence averages 901 frames, with a maximum of 3,858 missing frames.  
\textbf{RGBT234-Miss} covers all 234 sequences from RGBT234, totaling 116.7K frames, of which 69.2K frames simulate missing modalities, averaging 296 missing frames per sequence.  
\textbf{VTUAV-Miss} includes 176 test sequences from VTUAV, with 631.5K frames in total, 368.5K of which are missing, averaging 2,094 missing frames per sequence and a maximum of 4,097 missing frames.

\noindent\textit{Evaluation Protocol.} 
All experiments follow the one-pass evaluation (OPE) protocol, reporting precision rate (PR) and success rate (SR) as primary metrics in line with standard RGBT tracking evaluations.

\begin{table*}[t]
\centering
\caption{Comparison of SOTA trackers on RGBT datasets and missing-modality settings. 
The best results are highlighted in \textbf{bold red}, and the second-best performance is highlighted in \textbf{bold blue}. The “--” indicates absent results.}
\resizebox{1.0\textwidth}{!}{
\begin{tabular}{c|c|c|cc|cc|cc|cc|cc|cc}
\toprule
\multirow{2}{*}{Method} & \multirow{2}{*}{Pub.Info} & \multirow{2}{*}{Backbone} &
\multicolumn{2}{c|}{LasHeR} &
\multicolumn{2}{c|}{LasHeR-Miss} &
\multicolumn{2}{c|}{RGBT234} &
\multicolumn{2}{c|}{RGBT234-Miss} &
\multicolumn{2}{c|}{VTUAV} &
\multicolumn{2}{c}{VTUAV-Miss} \\
% \cmidrule(lr){3-14}
& & & PR$\uparrow$ & SR$\uparrow$ & PR$\uparrow$ & SR$\uparrow$
& MPR$\uparrow$ & MSR$\uparrow$ & MPR$\uparrow$ & MSR$\uparrow$
& PR$\uparrow$ & SR$\uparrow$ & PR$\uparrow$ & SR$\uparrow$ \\
\midrule
\textbf{SCDT} & - & ViT-B & \textcolor{red}{\textbf{77.4}} & \textcolor{blue}{\textbf{61.0}}& \textcolor{red}{\textbf{69.3}}& \textcolor{red}{\textbf{54.4}}& \textcolor{red}{\textbf{93.1}} & \textcolor{blue}{\textbf{69.6}} & \textcolor{red}{\textbf{88.1}} & \textcolor{red}{\textbf{64.3}} & \textcolor{red}{\textbf{93.6}} & \textcolor{red}{\textbf{78.9}} & \textcolor{red}{\textbf{84.1}} & \textcolor{red}{\textbf{69.6}} \\
FlexTrack~\cite{tan2025you} & ICCV 2025 & Fast-iTPN-B & \textcolor{blue}{\textbf{77.3}} & \textcolor{red}{\textbf{62.0}}& \textcolor{blue}{\textbf{65.1}} & \textcolor{blue}{\textbf{52.3}} & \textcolor{blue}{\textbf{92.7}} & \textcolor{red}{\textbf{69.9}} & \textcolor{blue}{\textbf{84.1}} & \textcolor{blue}{\textbf{62.6}} & -- & -- & -- & -- \\
IPL~\cite{lu2025modality} & IJCV 2025 & ViT-B & 69.4 & 55.3 & 61.7 & 46.4 & 88.3 & 65.7 & 82.0 & 59.4 & 87.5 & 75.6 & \textcolor{blue}{\textbf{80.9}} & \textcolor{blue}{\textbf{68.5}} \\
STTrack~\cite{hu2025exploiting} & AAAI 2025 & ViT-B & 76.0 & 60.3 & 54.5 & 44.9 & 89.8 & 66.7 & 73.8 & 54.2 & -- & -- & -- & -- \\
SUTrack~\cite{chen2025sutrack} & AAAI 2025 & HiViT-B & 74.5 & 59.5 & 58.3 & 47.6 & 92.2 & 69.5 & 82.0 & 60.8 & -- & -- & -- & -- \\
AINet~\cite{AINet2025} & AAAI 2025 & ViT-B & 74.2 & 59.1 & -- & -- & 89.2 & 67.3 & -- & -- & 88.0 & 75.3 & -- & -- \\
CAFormer~\cite{xiao2025cross} & AAAI 2025 & ViT-B & 70.0 & 55.6 & -- & -- & 88.3 & 66.4 & -- & -- & 88.6 & 76.2 & -- & -- \\
CKD~\cite{lu2024breaking} & ACM MM 2024 & ViT-B & 73.2 & 58.1 & -- & -- & 90.0 & 67.4 & -- & -- & \textcolor{blue}{\textbf{90.2}} & \textcolor{blue}{\textbf{77.8}} & -- & -- \\
OneTracker~\cite{hong2024onetracker} & CVPR 2024 & ViT-B & 67.2 & 53.8 & 50.0 & 39.9 & 85.7 & 64.8 & 70.8 & 49.9 & -- & -- & -- & -- \\
SDSTrack~\cite{hou2024sdstrack} & CVPR 2024 & ViT-B & 66.5 & 53.1 & 52.5 & 43.1 & 84.8 & 63.5 & 67.0 & 48.8 & -- & -- & -- & -- \\
UnTrack~\cite{wu2024single} & CVPR 2024 & ViT-B & 64.6 & 51.3 & -- & -- & 84.2 & 62.5 & -- & -- & -- & -- & -- & -- \\
ViPT~\cite{zhu2023visual} & CVPR 2023 & ViT-B & 65.1 & 52.5 & 44.0 & 36.9 & 83.5 & 61.7 & 52.4 & 39.4 & 85.0 & 73.0 & 63.0 & 54.5 \\
TBSI~\cite{hui2023bridging} & CVPR 2023 & ViT-B & 69.2 & 55.6 & 59.0 & 47.0 & 87.1 & 63.7 & 76.1 & 55.2 & -- & -- & -- & -- \\
% MCITrack~\cite{kang2025exploring} & AAAI 2025 & Fast-iTPN & 64.2 & 50.2 & 34.2 & 40.0 & 79.1 & 60.4 & 53.6 & 40.9 & -- & -- & -- & -- \\
ProTrack~\cite{yang2022prompting} & ACM MM 2022 & ViT-B & 53.8 & 42.0 & -- & -- & 79.5 & 59.9 & -- & -- & -- & -- & -- & -- \\
HMFT~\cite{Zhang_CVPR22_VTUAV} & CVPR 2022 & ResNet50 & -- & -- & -- & -- & 78.8 & 56.8 & -- & -- & 75.8 & 62.7 & 60.5 & 49.2 \\
OSTrack~\cite{ye2022joint} & ECCV 2022 & ViT-B & 51.5 & 41.2 & -- & -- & 72.9 & 54.9 & -- & -- & 85.0 & 73.4 & 63.3 & 54.5 \\
APFNet~\cite{xiao2022attribute} & AAAI 2022 & VGG-M & 50.0 & 36.2 & 38.2 & 29.0 & 82.7 & 57.9 & 68.4 & 47.0 & -- & -- & -- & -- \\
ADRNet~\cite{zhang2021learning} & IJCV 2021 & VGG-M & -- & -- & -- & -- & 80.7 & 57.0 & -- & -- & 62.2 & 46.6 & -- & -- \\
CAT~\cite{2020JMMAC} & ECCV 2020 & VGG-M & 45.0 & 31.4 & 30.4 & 22.5 & 80.4 & 56.1 & 57.5 & 39.6 & -- & -- & -- & -- \\
mfDiMP~\cite{zhang2019multi} & ICCVW 2019 & ResNet50 & 58.3 & 45.6 & 37.7 & 29.1 & 82.4 & 58.3 & 69.4 & 48.1 & 69.4 & 57.1 & 63.2 & 51.8 \\

\bottomrule
\end{tabular}}
\label{tab:table1}
\end{table*}

% CMPP~\cite{cmpp} & -- & -- & -- & -- & 82.3 & 57.5 & -- & -- \\
% JMMAC~\cite{jmmac} & -- & -- & -- & -- & 79.0 & 57.3 & -- & -- \\

% FANet~\cite{fanet} & 44.1 & 30.9 & -- & -- & 78.7 & 55.3 & -- & -- \\
% mfDiMP~\cite{mfdimp} & 44.7 & 34.3 & -- & -- & 64.6 & 42.8 & -- & -- \\
% SGT~\cite{sgt} & 36.5 & 25.1 & -- & -- & 72.0 & 47.2 & -- & -- \\
% HMFT~\cite{hmft} & 43.6 & 31.3 & -- & -- & -- & -- & -- & -- \\
% DAPNet~\cite{dapnet} & 43.1 & 31.4 & -- & -- & -- & -- & -- & -- \\
% DAFNet~\cite{dafnet} & -- & -- & -- & -- & 79.6 & 54.4 & -- & -- \\
% MaCNet~\cite{macnet} & -- & -- & -- & -- & 79.0 & 55.4 & -- & -- \\

\subsection{Evaluations on complete dataset}
We evaluate SCDT on three widely used complete-modality benchmarks: LasHeR, RGBT234, and VTUAV. As shown in Table~\ref{tab:table1}, SCDT achieves competitive or state-of-the-art performance across all datasets.
On LasHeR, SCDT achieves the best PR score of 77.4\% and the second-best SR score of 61.0\%, slightly surpassing the previous leading method FlexTrack in PR. On RGBT234, SCDT sets a new state-of-the-art MPR of 93.1\% and obtains the second-best MSR of 69.6\%, outperforming most recent ViT-based trackers by clear margins. On the VTUAV, SCDT also achieves the highest PR/SR scores of 93.6\%/78.9\%.
Overall, SCDT demonstrates strong effectiveness under complete-modality conditions, confirming that its conditional enhancement mechanism improves cross-modal feature quality even without modality degradation.

\subsection{Evaluations on missing RGBT dataset}
To assess robustness under modality-missing, we further evaluate SCDT on three challenging missing-modality benchmarks: LasHeR-Miss, RGBT234-Miss, and VTUAV-Miss. Each benchmark includes multiple missing patterns and ratios, providing a comprehensive test bed for reconstruction-based multimodal tracking.
As summarized in Table~\ref{tab:table1}, SCDT achieves clear state-of-the-art performance across all three datasets. On LasHeR-Miss, SCDT reaches 69.3\% PR and 54.4\% SR, outperforming the previous best FlexTrack by 4.2\% and 2.1\%, respectively. On RGBT234-Miss, SCDT establishes new SOTA scores of 88.1\% MPR and 64.3\% MSR, significantly surpassing recent methods, including FlexTrack and IPL. On VTUAV-Miss, SCDT again achieves the best PR/SR results of 84.1\%/69.6\%.
These results demonstrate the strong capability of SCDT to handle incomplete multimodal inputs. The 
% proposed spatio-temporal conditioned 
denoising mechanism enables effective reconstruction of missing information and maintains robust performance across diverse missing scenarios.

% Recently, modality missing object tracking has started attracting increasing attention, as real-world multimodal systems often encounter incomplete inputs where one or more modalities are missing. We evaluate our method on three missing RGBT tracking datasets, including LasHeR-Miss, RGBT234-Miss and VTUAV-Miss. These benchmarks contains various missing patterns with combining five missing patterns with three missing ratio, covering all real-world modality missing scenarios.
% %

% As shown in Table 1, SCDT demonstrates outstanding performance on the LasHeR-Miss~\cite{li2021lasher}, RGBT234-Miss~\cite{li2019rgb234}, and VTUAV-Miss~\cite{Zhang_CVPR22_VTUAV} datasets. Specifically, on the LasHeR-Miss benchmark, SCDT achieves PR score of 69.3\% and an AUC score of 54.4\%, surpassing the previous best-performing model, FlexTrack, by 4.2\% and 2.1\%, respectively. On the RGBT234-Miss benchmark, SCDT sets new SOTA scores for MPR at 88.1\% and MSR at 64.3\%, showing a advantage of 4.0\%/1.7\% and 6.1\%/3.9\% in PR/SR compared to FlexTrack and IPL, respectively. On the VTUAV-Miss benchmark, we also focus on the short-term tracking subset. SCDT achieves state-of-art performance of 84.1\% and 69.6\% on PR and SR scores.
% %
% These results indicate that with Spatio-temporal guidance, SCDT significantly improves performance under missing modality conditions, enabling its ability to effectively handle missing modalities and maintain robustness in challenging real-world tracking scenarios.

\subsection{Ablation Study}
This section investigates the contributions of key components in SCDT, including temporal conditioning, noise-modulated adaptation, task-specific supervision and denoising depth. All experiments are conducted on standard RGBT tracking benchmarks, with precision rate (PR), mean precision rate (MPR), success rate (SR) and mean success rate (MSR) as primary evaluation metrics.

\begin{table}[t]
\centering
\caption{Ablation studies on Spatio-temporal conditions.}
\label{tab:ablation_denoising_conditions}
\setlength{\tabcolsep}{2.5mm}
\renewcommand{\arraystretch}{1}{
\resizebox{\linewidth}{!}{
\begin{tabular}{l|c c|c c|c c|c c}
\toprule
 \multirow{2}{*}{Method} & \multicolumn{2}{c|}{LasHeR} & \multicolumn{2}{c|}{LasHeR-Miss}  & \multicolumn{2}{c|}{RGBT234} & \multicolumn{2}{c}{RGBT234-Miss}\\
 & PR & SR & PR & SR  & MPR & MSR & MPR & MSR\\
\midrule
 baseline & 75.1&59.2 & 63.2&49.6& 91.6&69.2 & 83.8&60.8 \\ 
 % baseline-missep30-random & 75.4 & 67.2\\
 %+ spatial conditionv1  & 75.7 & 66.9 & 92.9/68.5& 84.6/59.7\\
 w/ SP  & 75.7&58.7 & 66.9&52.2 & 92.9&67.5& 85.8&60.9\\
 % + temporal condition  & 75.2 & 68.3\\  & 75.9 & 68.3\\ 
 w/ SP ST & 75.8&59.6 & 68.3&53.6& 92.6&68.4&86.1&62.2\\
 w/ SP LT &  76.0&59.7 & 67.3&52.9 & 92.9&69.3&86.2&62.4\\
 %+ spatial \& short-term temporalv1 & 75.8 & 68.3& 92.4/68.5&87.2/62.9\\
\midrule
 %+ spatial \& short-time condition-learnratev0 & 76.0 & 68.7\\
 Ours & \textbf{77.4} &\textbf{61.0} & \textbf{69.3} & \textbf{54.4} & \textbf{93.1} & \textbf{69.6} & \textbf{88.1} &\textbf{64.3}\\
\bottomrule
\end{tabular}}}
\end{table}

\noindent\textbf{Impact of Temporal Conditions.} %验证短期与长期时序条件在缺失模态重建和完整模态增强中的作用。
\begin{table}[t]
\centering
\small
%\fontsize{6}{8}\selectfont % 字体大小设为8pt，行距为10pt
\caption{Ablation studies on Noise-Modulated Adaptation.}
\label{tab:tab_noise}
\begin{tabular}{c|c c|c c}
\toprule
\multirow{2}{*}{Complete - Missing}& \multicolumn{2}{c|}{LasHeR} & \multicolumn{2}{c}{LasHeR-Miss} \\
 & PR & SR & PR & SR \\
\midrule
strong - strong & 73.2 & 57.4 & 65.4 & 51.4 \\
weak - weak & 75.8 & 59.6 & 68.9 & 54.1\\
weak - strong & \textbf{77.4} & \textbf{61.0} & \textbf{69.3} & \textbf{54.4} \\
\bottomrule
\end{tabular}
\end{table}
We first analyze the role of different temporal conditioning strategies in the SCDT module. As shown in Table~\ref{tab:ablation_denoising_conditions}, three types of conditioning are conducted: spatial-only (w/ SP), spatial with short-term temporal cues (w/ SP ST), and spatial with long-term temporal cues (w/ SP LT).
Adding spatial condition (w/ SP) already improves reconstruction quality under missing modalities, verifying that the conditional denoiser benefits from explicit spatial guidance. Introducing short-term temporal cues (w/ SP ST) yields further gains, particularly under modality degradation. On LasHeR-Miss, short-term conditioning boosts PR/SR by +1.4\%/+1.4\% over w/ SP, reflecting its ability to enforce fine-grained motion continuity and stabilize reconstruction against local temporal fluctuations.
Long-term condition (w/ SP LT) instead enhances performance on complete-modality benchmarks. On LasHeR and RGBT234, long-term cues improve PR/SR by +0.3\%/+0.1\% and MPR/MSR by +0.3\%/+0.9\% over w/ SP ST, indicating that global temporal context strengthens high-level semantic consistency and reduces accumulated drift.
When short-term and long-term cues are combined, SCDT consistently achieves the best overall performance. Compared with the strongest single-condition variant, the full model further improves LasHeR by +1.4\% PR and +1.3\% SR, and LasHeR-Miss by +1.0\%/+0.8\% PR/SR. These gains demonstrate that multi-scale temporal conditioning is crucial for maintaining consistent cross-modal feature generation and ensuring stable RGBT tracking under both complete and missing conditions.

\noindent\textbf{Impact of Noise-Modulated Adaptation.}
We evaluate three noise strategies under complete and missing modality scenarios, as reported in Table~\ref{tab:tab_noise}. The strategies are: applying strong noise to both scenarios, applying weak noise to both, and applying weak noise to complete modalities while strong noise is applied to missing modalities, which is the design used in SCDT.
Applying strong noise to both scenarios yields limited performance because the alignment loss cannot generate high-quality features from heavily corrupted inputs. Using weak noise for both improves results by preserving feature integrity, but it provides insufficient guidance to reconstruct missing modalities. 
The weak-to-strong configuration achieves the best performance. Weak noise introduces mild perturbation to complete modalities, improving robustness, while strong noise simulates missing modalities, providing effective reconstruction guidance. This strategy balances enhancement and reconstruction, allowing a single model to handle both complete and missing modalities effectively.

\begin{table}[t]
\centering
\small
\caption{Ablation studies under different supervision.}
\label{tab:loss_table}
\begin{tabular}{l|c c|c c}
\toprule
 \multirow{2}{*}{Supervisions} & \multicolumn{2}{c|}{LasHeR} & \multicolumn{2}{c}{LasHeR-Miss} \\
 & PR & SR & PR & SR \\
\midrule
$\mathcal{L}_{\text{align}}$ & 76.0 & 59.9 & 67.2 & 52.8 \\
$\mathcal{L}_{\text{recon}}$ & 75.6 & 59.5 & 68.0 & 53.3 \\ 
$\mathcal{L}_{\text{align}}$ + $\mathcal{L}_{\text{recon}}$ & \textbf{77.4} & \textbf{61.0} & \textbf{69.3} & \textbf{54.4} \\
\bottomrule
\end{tabular}
\end{table}

\noindent\textbf{Impact of Task-Specific Supervision.}
We investigate the effect of different supervision strategies on feature generation under complete and missing modality scenarios. Table~\ref{tab:loss_table} compares using the alignment loss alone, the reconstruction loss alone, and their combination. Employing only the alignment loss achieves competitive performance on complete scenes but underperforms on missing modality sequences, as it lacks explicit reconstruction guidance. Conversely, using only the reconstruction loss benefits missing modality recovery but does not fully enhance features for complete modalities. Combining alignment and reconstruction losses delivers the best results, with LasHeR achieving 77.4\% PR and 61.0\% SR and LasHeR-Miss reaching 69.3\% PR and 54.3\% SR. This demonstrates that task-specific supervision effectively balances feature enhancement and reconstruction, producing stable and discriminative representations under both complete and missing modality conditions.

\begin{table}[t]
\centering
\small
\caption{Ablation studies on the depth of denoising layers.}
\label{tab:depth_tab}
\begin{tabular}{c|c c|c c}
\toprule
  \multirow{2}{*}{Layers number}& \multicolumn{2}{c|}{LasHeR} & \multicolumn{2}{c}{LasHeR-Miss} \\
 & PR & SR & PR & SR \\
\midrule
2  & 75.9 & 59.0 & 68.8 & 54.0 \\
4  & \textbf{77.4} & \textbf{61.0} & \textbf{69.3} & \textbf{54.4}\\
6  & 76.4 & 60.0 & 69.2 & 54.3 \\
\bottomrule
\end{tabular}
\end{table}

\label{sec:formatting}
\begin{figure}
  \centering
    %\fbox{\rule{0pt}{2in} \rule{.9\linewidth}{0pt}}
    \includegraphics[width=0.9\linewidth]{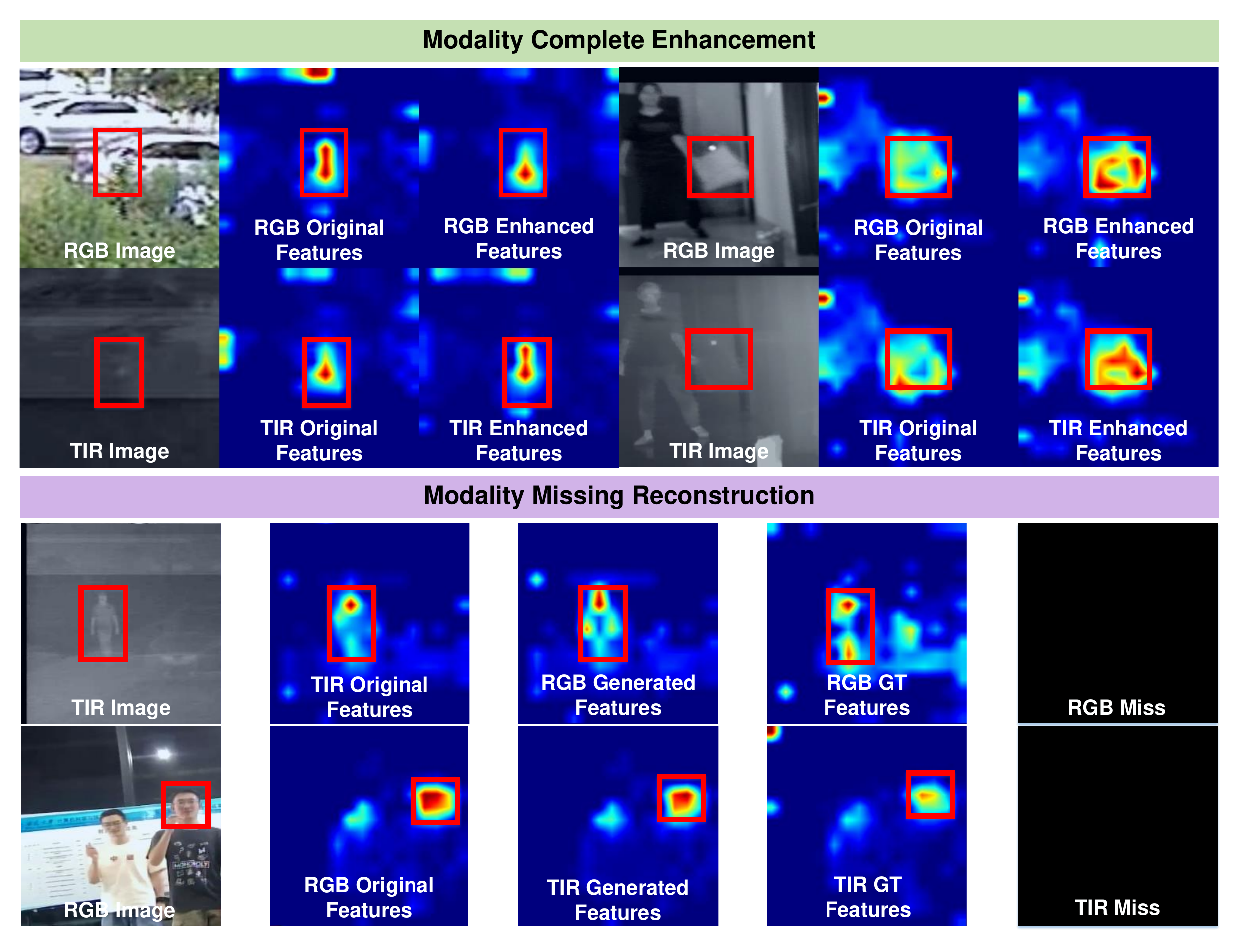}
    \caption{Visualization of features learned by SCDT under complete and missing modality conditions.}
    \label{fig:feature_visv3}
\end{figure}

\begin{figure*}[t]
    \centering
    \includegraphics[width=0.9\linewidth]{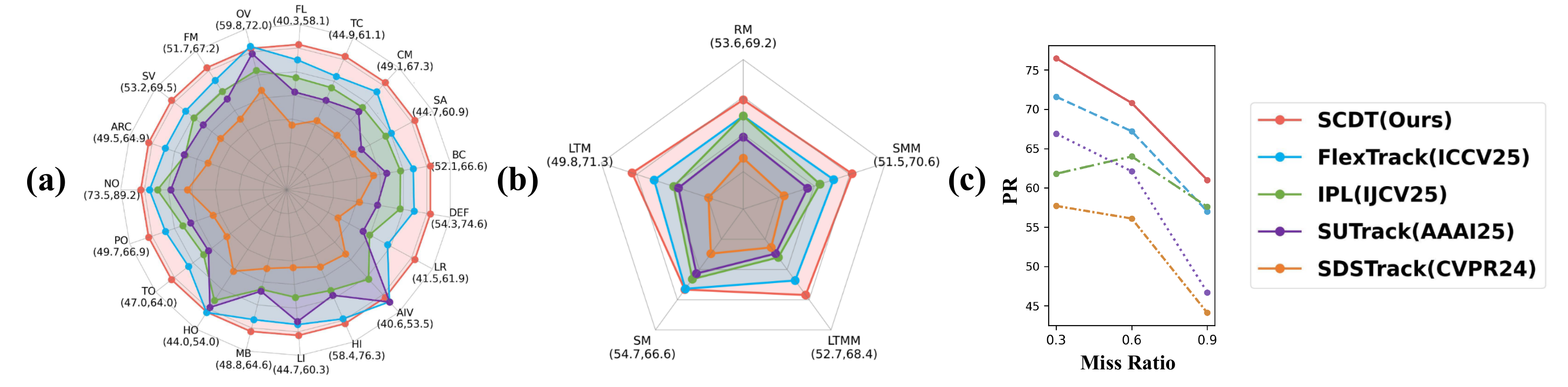}
     \caption{Precision rate (PR) of challenge attributes on LasHeR-Miss dataset. The axes of each attribute have been normalized.
     }
    \label{Challenges}
\end{figure*}

\noindent\textbf{Impact of Denoising Layer Depth.}
Table~\ref{tab:depth_tab} shows that the depth of the denoising module plays a critical role in refinement quality. A shallow two-layer design yields weaker feature correction, leading to reductions of 1.5\% PR on LasHeR and 0.5\% PR on LasHeR-Miss. Increasing the depth to six layers does not bring further benefits and slightly suppresses performance due to redundant refinement and over-smoothing. The four-layer configuration achieves the strongest results, reaching 77.4\% PR and 61.0\% SR on LasHeR and 69.3\% PR and 54.4\% SR on LasHeR-Miss. These results indicate that a moderate depth provides the most effective balance between denoising capacity and stable feature enhancement.

% \noindent\textbf{Effectiveness of the Denoising Formulation.}
% \textcolor{red}{To demonstrate that our approach is essential rather than merely a specific training recipe for feature reconstruction and fusion, we also compare our denoiser against a standard regression-based reconstruction network. As shown in Table~\ref{tab:reg_de}, replacing the denoising formulation with one-shot regression leads to a noticeable performance drop. This decline occurs because regression produces over-smoothed features. In contrast, our iterative denoising explicitly suppresses background false responses and improves overall robustness."}

\subsection{Qualitative Analysis}

\noindent\textbf{Feature Visualization.} To qualitatively verify the effectiveness of SCDT, we visualize the learned representations under complete and missing modality conditions in Fig.~\ref{fig:feature_visv3}. In the modality-complete case, SCDT produces feature maps that are sharply concentrated on the target while effectively suppressing background responses. This demonstrates its ability to refine weak modalities and strengthen discriminative cues through spatio-temporal conditioning.  
In the modality-missing case, the generated features closely match the ground-truth features of the missing modality in both spatial alignment and activation structure, indicating high-fidelity reconstruction.  
These visualizations confirm that noise-modulated adaptation and dual conditioning jointly enable SCDT to enhance reliable modalities and reconstruct missing ones, yielding stable and coherent multimodal features regardless of modality availability.

\noindent\textbf{Attribute-wise Robustness Analysis.}
We evaluate SCDT across a wide range of challenging attributes, and the results are presented in Fig.~\ref{Challenges}.

\noindent\textit{Conventional Tracking Challenges.}
Fig.~\ref{Challenges} (a) shows performance on the LasHeR-Miss benchmark across all attributes, SCDT consistently outperforms prior methods on No Occlusion (NO), Partial Occlusion (PO), Total Occlusion (TO), Hyaline Occlusion (HO), Motion Blur (MB), Low Illumination (LI), High Illumination (HI), Low Resolution (LR), Deformation (DEF), Background Clutter (BC), Similar Appearance (SA), Camera Moving (CM), Thermal Crossover (TC), Frame Lost (FL), Out of View (OV), Fast Motion (FM), Scale Variation (SV), and Aspect Ratio Change (ARC). The largest gains appear in LR, SA, and TC, indicating that SCDT generates more stable and discriminative features under severe appearance degradation.

\noindent\textit{Modality-Missing Challenges.}
To further assess robustness under modality missing, we evaluate performance across different missing challenges shown in Fig.~\ref{Challenges} (b) and Fig.~\ref{Challenges} (c). These include missing-type patterns such as Long Time Missing (LTM), Switch Missing (SM), Random Missing (RM), Long Time Mixed Missing (LTMM), and Switch Mixed Missing (SMM), as well as missing ratios of 30\%, 60\%, and 90\%. SCDT consistently outperforms existing trackers in every pattern and ratio, with the most significant advantages under long-term missing and high-ratio settings. This demonstrates that the proposed noise-modulated adaptation and conditional denoising mechanism effectively stabilizes feature generation even when large portions of a modality are unavailable.
Overall, SCDT delivers uniform and substantial gains across all conventional and modality-missing challenges, confirming its strong resilience to visual challenges and modality missing.

\begin{figure}[t]
    \centering
    \includegraphics[width=0.9\linewidth]{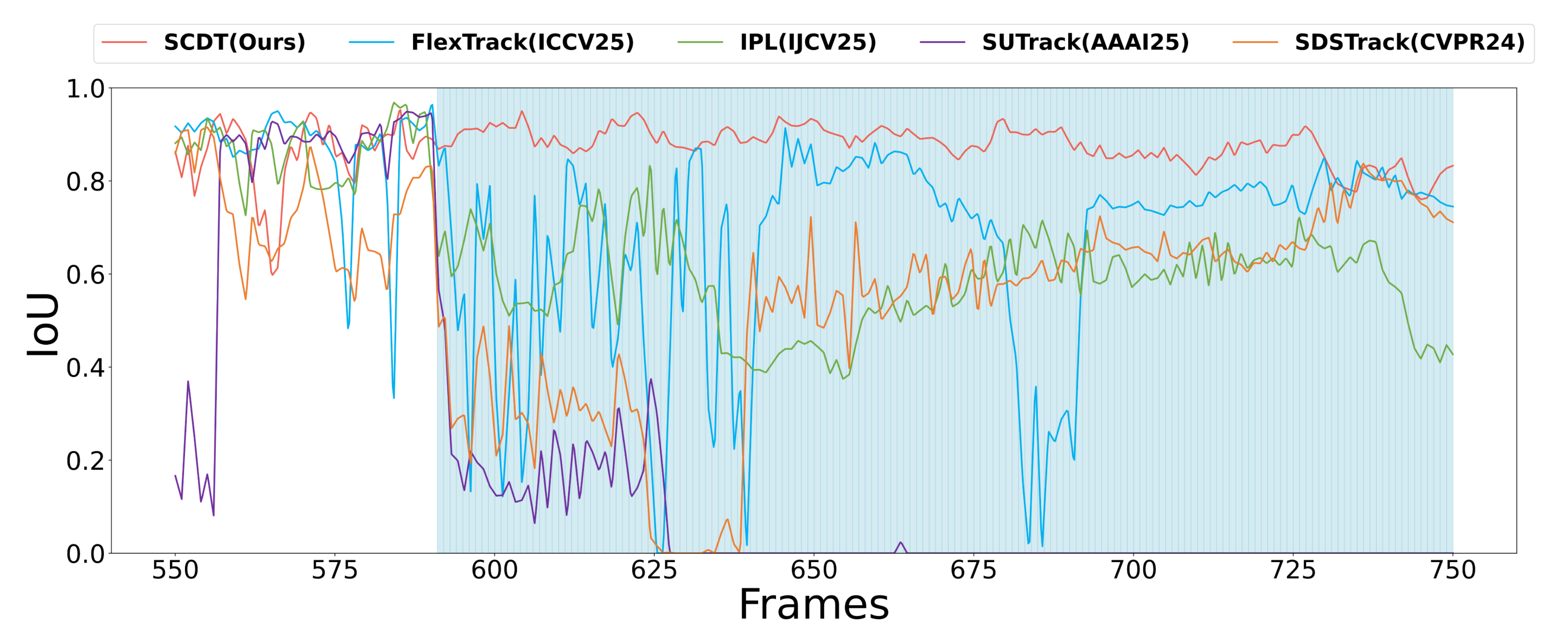}
    \caption{Per-frame tracking IoU curves in the leftmirror sequence under modality-missing challenges. Blue shaded regions indicate frames from RGB modality is missing.}
    \label{iou}
\end{figure}

\noindent\textbf{Tracking Visualization.} 
We present per-frame tracking IoU curves under modality-missing challenges in Fig.~\ref{iou}, comparing SCDT with both modality-missing and conventional state-of-the-art trackers. Although most methods perform similarly at the start of sequences, our method shows clear advantages on frames with missing modalities.

\section{Conclusion}

%-------------------------------------------------------------------------
In this paper, we have proposed the Spatio-temporal Conditional Denoising Transformer (SCDT) for robust RGBT tracking under both complete and missing modality conditions. By integrating temporal-conditioned feature generation, noise-modulated adaptation, and task-specific supervision, SCDT enables a single model to enhance reliable modalities while reconstructing missing ones, avoiding separate designs or modality-specific heuristics. Extensive experiments show consistent gains across modality-complete and modality-missing benchmarks, demonstrating the effectiveness of conditioned denoising in producing resilient multimodal representations. Overall, this framework reveals conditioned denoising as a promising direction for robust multimodal perception, offering a scalable foundation for future research in flexible modality handling, generative fusion, and real-world multimodal tracking.
\section{Acknowledgments}
This work was supported in part by the National Natural Science Foundation of China (No. 62406002 and No. 62376004); and the Anhui Province Science Fundation for Distinguished Young Scholars (No. 2208085J18).

% \noindent\textbf{Limitations.} 
% While SCDT demonstrates strong performance across diverse RGBT tracking scenarios, its current design primarily focuses on two modalities. Extending the framework to handle additional sensors or heterogeneous inputs could further enhance flexibility. Moreover, although the model generalizes well under simulated modality-missing conditions, exploring real-world sensor failures and asynchronous frame rates may provide additional insights for deployment. 

% These directions offer natural extensions to strengthen the robustness and applicability of spatio-temporal conditional denoising for broader multimodal perception tasks.

% Our framework introduces a novel approach to multimodal fusion, where the generation and enhancement of features are dynamically guided by noise-conditioned spatiotemporal priors. This design ensures that our model can seamlessly adapt to varying input conditions, effectively bridging gaps in missing modalities while enriching the representation in complete-modality cases. Extensive experiments on six benchmark datasets demonstrate that SCDT not only outperforms state-of-the-art methods across a wide range of tracking scenarios, but also offers a flexible and scalable solution for RGBT tracking in dynamic and complex environments.

%-------------------------------------------------------------------------
{
    \small
    \bibliographystyle{ieeenat_fullname}
    \bibliography{main}
}

%     \small
%     \bibliographystyle{ieeenat_fullname}
%     \bibliography{suppl}
% }

\end{document}